%% file: main.tex
\documentclass[sigconf, arxiv,nonacm]{acmart} 
\usepackage{dsfont}
\usepackage{subfigure}

\AtBeginDocument{%
  \providecommand\BibTeX{{%
    \normalfont B\kern-0.5em{\scshape i\kern-0.25em b}\kern-0.8em\TeX}}}

\begin{document}

\title{Simple Truncated SVD based Model for Node Classification on Heterophilic Graphs}

\author{Vijay Lingam, Rahul Ragesh, Arun Iyer, Sundararajan Sellamanickam}

\email{t-vili, t-rarage, ariy, ssrajan@microsoft.com}

\affiliation{%
  \institution{Microsoft Research India}
  \city{Bengaluru, India}
}

\input{commands}
\input{sections/abstract}
\maketitle

\input{sections/introduction}

\input{sections/relatedworks}
\input{sections/approach}

\input{sections/experiments}

\input{sections/conclusion}

\bibliographystyle{ACM-Reference-Format}
\vspace{30mm}
\bibliography{references}

\end{document}

%% file: commands.tex
\newcommand{\graph}{\mathcal{G}}
\newcommand{\adj}{\mathbf{A}}
\newcommand{\verts}{\mathcal{V}}
\newcommand{\edges}{\mathcal{E}}
\newcommand{\train}{T}
\newcommand{\dadj}{\mathbf{D}_{\adj}}
\newcommand{\adji}{\mathbf{A_I}}
\newcommand{\dadji}{\mathbf{D}_{\adji}}
\newcommand{\feat}{\mathbf{X}}
\newcommand{\eye}{\mathbf{I}}
\newcommand{\real}{\mathbb{R}}
\newcommand{\eigenU}{\mathbf{U}}
\newcommand{\eigenUc}{\mathbf{u}}
\newcommand{\adjS}{\mathbf{\Sigma}}
\newcommand{\adjSv}{\sigma}
\newcommand{\eigenV}{\mathbf{V}}
\newcommand{\adjU}{\eigenU_{\adji}}
\newcommand{\adjV}{\eigenV_{\adji}}
\newcommand{\featU}{\eigenU_{\feat}}
\newcommand{\featV}{\eigenV_{\feat}}

\newcommand{\gcn}{\textsc{GCN}}
\newcommand{\sgcn}{\textsc{SGCN}}
\newcommand{\geomgcn}{\textsc{Geom-GCN}}
\newcommand{\supergat}{\textsc{SuperGAT}}
\newcommand{\hhgcn}{\textsc{H\textsubscript{2}GCN}}
\newcommand{\fagcn}{\textsc{FAGCN}}
\newcommand{\gprgnn}{\textsc{GPR-GNN}}
\newcommand{\eigennet}{\textsc{EigenNetwork}}
\newcommand{\eigenconcat}{\textsc{Eigen-ConcatNetwork}}
\newcommand{\eigeneigennet}{\textsc{Eigen-EigenNetwork}}
\newcommand{\tiedeigen}{\textsc{RegEigenNetwork}}
\newcommand{\appnp}{\textsc{APPNP}}
\newcommand{\lgcn}{\textsc{LightGCN}}
\newcommand{\hetegcn}{\textsc{HeteGCN}}


%% file: sections/abstract.tex
\begin{abstract}
Graph Neural Networks (GNNs) have shown excellent performance on graphs that exhibit strong homophily with respect to the node labels i.e. connected nodes have same labels. However, they perform poorly on heterophilic graphs. Recent approaches have typically modified aggregation schemes, designed adaptive graph filters, etc. to address this limitation. In spite of this, the performance on heterophilic graphs can still be poor. We propose a simple alternative method that exploits Truncated Singular Value Decomposition (TSVD) of topological structure and node features. Our approach  achieves up to $\sim$30\% improvement in performance over state-of-the-art methods on heterophilic graphs. This work is an early investigation into methods that differ from aggregation based approaches. Our experimental results suggest that it might be important to explore other alternatives to aggregation methods for heterophilic setting. 
\end{abstract}

%% file: sections/introduction.tex
\section{Introduction}
Homophily~\citep{homophily} is a principle in sociology that suggests that connections in real life are bred through similarity. In the context of semi-supervised classification, this implies that nodes with similar labels are likely to be connected. Several real world networks exhibit homophily, for example, people on a social network connect to each other based on similar interests. There are also several real world networks that exhibit the opposite behaviour. For example, the Wikipedia page on Homophily is not only connected to other pages from sociology, but also connected to various pages from mathematics, graph theory and statistics. Since Wikipedia is a large body of collective knowledge, its pages often have connections between several different areas.

Graph Neural Networks (GNNs)~\citep{gcn, graphsage, gat} leverage network information along with node features to improve their semi-supervised classification performance. GNNs are largely dependent on network homophily to be able to give improved performance. For heterophilic networks, their performance can degrade significantly. Several approaches have been proposed in the literature to mitigate this degradation in performance in presence of heterophily. These approaches can be organized into three groups. The first of these approaches involve modifying the aggregation mechanism in graph neural networks in an effort to mitigate issues caused by heterophily. \citet{geomgcn} proposed message passing both over the graph neighborhood and the neighbors in the latent space. ~\citet{h2gcn} proposed to keep the self embedding separate from the neighbor embeddings during aggregation, while also incorporating higher order neighbor embeddings in a similar fashion. ~\citet{supergat} proposed several simple attention models trained on an additional auxiliary task and finally present an analysis of which attention model is well suited for homophily and heterophily. The second way to address heterophily is to explicity model a label-label compatibility matrix learnt that can be used as a prior to update the posterior belief in the label predictions. \citet{cpgnn} proposes to model the label compatibility matrix that reflects the heterophily in the graph and utilizes the model in a GNN. The third group of approaches involve designing graph filters that can directly adapt to low frequency as well as high frequency parts of the graph as needed by the model. \citet{fagcn} proposes to learn an attention mechanism that captures the proportion of low-frequency and high-frequency signals per edge. \citet{gprgnn} proposes an adaptive polynomial filter to pick up which low-frequency high-frequency signals are helpful for the task.

For homophilic networks, existing models \citet{gprgnn, appnp} already prove to be excellent. Our focus lies in heterophilic networks. We are interested in class of methods that aim at modifying or adapting the graph to obtain better performance in heterophilic graphs~\cite{supergat, fagcn, gprgnn}. The more recent approaches among these class of methods adjust the eigenvalues of the graph to learn improved representations. Another interpretation for this adaptation is that it is selecting eigenvectors by learning coefficients. We replace this complex adaptation by a TSVD and propose simple yet effective methods to improve task performance. Our contributions can be summarized as follows.
\begin{enumerate}
    \item We present simple TSVD based methods based on our insights that outperform state-of-the-art approaches on heterophilic networks with performance gains of up to $\sim$ 30\%.
    \item Deviating from the popular message-passing frameworks, we propose a simple and efficient concatenation-based model that is competitive to neighborhood aggregation-based models and outperforms several baselines on benchmark datasets.
\end{enumerate}
In the following sections, we discuss related works, motivate our proposed approach, and finally present our experimental results along with ablative studies.

%% file: sections/relatedworks.tex
\section{Related Works}
\label{sec:relatedworks}

Most of the development in the GNNs were for homophilic graphs, and they performed poorly in heterophily setting. One of the early works to address heterophily in GNNs was \geomgcn~\citep{geomgcn}. They identified two key weaknesses in GNNs in the context of heterophily. First, since the aggregation over the neighbourhood is permutation-invariant, it is difficult to identify which neighbours contribute positively and negatively to the final performance. Second, long-range information is difficult to aggregate. To mitigate these issues, they proposed aggregating over two sets of the neighbourhood - one from the graph and the other inferred in the latent space. \hhgcn~\citep{h2gcn} proposed to separate the self-embeddings from neighbour embeddings. To avoid mixing of information, they concatenate self-embeddings and neighbour embeddings instead of aggregating them. Higher-order neighbourhood embeddings are similarly combined to capture long-range information. 

Recent approaches address these shortcomings by adapting the graph. \supergat~\citep{supergat} gave several simple attention models trained on the classification and an additional auxiliary task. They suggest that these attention models can improve model performance across several graphs with varying homophily scores. \fagcn~\citep{fagcn} uses the attention mechanism and learns the weight of an edge as the difference in the proportion of low-frequency and high-frequency signals. They empirically show that negative edge-weights identify edges that connect nodes with different labels. \gprgnn~\citep{gprgnn} takes the idea proposed in \appnp~and generalizes the Pagerank model that works well for graphs with varying homophily scores. Our proposed approach greatly simplifies adaptation methods while delivering significant performance improvements for the task at hand. 

%% file: sections/approach.tex
\section{Proposed Approach}
\renewcommand{\arraystretch}{1.2}

\begin{table*}[htp]
\centering
\resizebox{\textwidth}{!}{

\begin{tabular}{ccccccccc}
\hline
\textbf{Dataset} &
  \textbf{Texas} &
  \textbf{Wisconsin} &
  \textbf{Actor} &
  \textbf{Squirrel} &
  \textbf{Chameleon} &
  \textbf{Crocodile} &
  \textbf{Cornell} \\ \hline
\multicolumn{1}{c|}{\textbf{Homophily level}} & 0.11 & 0.21 & 0.22  & 0.22   & 0.23  & 0.26   & 0.30 \\
\multicolumn{1}{c|}{\textbf{\#Nodes}}         & 183  & 251  & 7600  & 5201   & 2277  & 11631  & 183  \\
\multicolumn{1}{c|}{\textbf{\#Edges}}         & 492  & 750  & 37256 & 222134 & 38328 & 191506 & 478  \\
\multicolumn{1}{c|}{\textbf{\#Features}}      & 1703 & 1703 & 932   & 2089   & 500   & 500    & 1703  \\
\multicolumn{1}{c|}{\textbf{\#Classes}}       & 5    & 5    & 5     & 5      & 5     & 6      & 5     \\
\multicolumn{1}{c|}{\textbf{\#Train/Val/Test}} &
  87/59/37 &
  120/80/51 &
  3648/2432/1520 &
  2496/1664/1041 &
  1092/729/456 &
  120/180/11331 &
  87/59/37  \\ \hline
\end{tabular}

}
\vspace{1mm}
\caption{Datasets Statistics}
\vspace{-5mm}
\label{tab:stats_table}
\end{table*}

\subsection{Preliminaries}
We focus on the problem of semi-supervised node classification on a simple graph $\graph = (\verts, \edges)$, where $\verts$ is the set of vertices and $\edges$ is the set of edges. Let $\adj \in \{0, 1\}^{n \times n}$ be the adjacency matrix associated with $\graph$, where $n = |\verts|$ is the number of nodes. Let $\mathcal{Y}$ be the set of all possible class labels. Let $\feat \in \real^{n \times d}$ be the $d$-dimensional feature matrix for all the nodes in the graph. Given a training set of nodes $\train \subset \verts$ whose labels are known, along with $\adj$ and $\feat$, our goal is to predict the labels of the remaining nodes. The proportion of edges that connect two nodes with the same labels in a graph is called the homophily score of the graph. In our problem, we are particularly concerned with graphs that exhibit low homophily scores. In the next sub-section, we provide background material on the GPR-GNN modelling method and its approach to graph adaptation.

\label{sec:proposedapproach}
\subsection{\gprgnn~Model}  
The \gprgnn~\citep{gprgnn} model consists of two core components: (a) a non-linear network that transforms raw feature input $\feat$: ${\bf Z}^{(0)} = f(\feat;{\mathbf W})$ and (b) a generalized page ranking (GPR) component, ${\mathbf G}$, that essentially aggregates  the transformed output ${\mathbf Z}$ recursively as: ${\bf Z}^{(k)} = \adj{\bf Z}^{(k-1)}, k = 1, \ldots, K$. Notice that there is no nonlinear operation involved after each aggregation step over $k$. Therefore, the functionality of the GPR component can be written using an operator ${\mathbf G}$ defined as: ${\mathbf G} = \sum_{k=0}^K \alpha_k {\mathbf A}^k$
and we obtain aggregated node embedding by applying ${\mathbf G}$ on the nonlinear network output:  ${\mathbf S} = {\mathbf G} {\mathbf Z}^{(0)}$. 
Using singular value decomposition, ${\adj} = \eigenU \adjS \eigenU^T$, \citet{gprgnn} presented an interpretation that the GPR component essentially performs a graph filtering operation: ${\mathbf G} = \eigenU h_K(\adjS) \eigenU^T$ where $h_K(\adjS)$ is a polynomial graph filter applied element-wise and $h_K(\adjSv_i) = \sum_{k=0}^K \alpha_k \adjSv^k_i$ where $\adjSv_i$ is the $i^{th}$ eigen value. As explained in~\citet{gprgnn}, learning filter coefficients (i.e., $\alpha$) help to get improved performance. Since the coefficients can take negative values the \gprgnn~model is able to capture high-frequency components of the graph signals, enabling the model to achieve improved performance on heterophilic graphs.

\subsection{Proposed Approach}
In this section, we present an alternative interpretation of the \gprgnn~model and suggest a simple Truncated Singular Value Decomposition (TSVD) based method. We present two approaches, each motivated by considering different aspects of the problem. 

We start by closely observing the GPR component output given by: 
\begin{equation}
    {\mathbf S} = \eigenU h_K(\adjS) \eigenU^T {\mathbf Z}^{(0)}.
\label{eqn:GPRGNN2}    
\end{equation}
Our first observation is that learning filter coefficients is equivalent to learning a \textit{new} graph, ${\tilde \adj(\eigenU, \adjS;\alpha})$ which is dependent on the fixed set of singular vectors and singular values, but, parameterised using $\alpha$. Therefore, the \gprgnn~model may be interpreted as adapting the original adjacency matrix $\adj$. Next, as noted in the previous section, using the structures present in singular value decomposition and polynomial function, we can expand (\ref{eqn:GPRGNN2}) by unrolling over eigenvalues and interchanging the summation as:
\begin{equation}
    {\mathbf S} = \sum_{j=1}^n  h_K(\adjSv_j;\alpha) \eigenUc_j \eigenUc^T_j {\mathbf Z}^{(0)}.
\label{eqn:GPRGNN3}    
\end{equation}
There are several choices available for selection functions that one can choose from. However, we simplify $h_K(\adjSv_j; \alpha)$ by replacing it with a simple selection function as follows:
\begin{equation}
    {\mathbf S} = \sum_{j=1}^n  \mathds{1}_{j \leq k_{1}} \adjSv_j \eigenUc_j  \eigenUc^T_j {\mathbf Z}^{(0)}.
\label{eqn:truncatedEigen}    
\end{equation}


Our selection function is equivalent to a TSVD, where $k_{1}$ largest singular values are used for reconstructing the new graph. Restricting to $k_{1}$ singular values naturally induces negative edges. It is worth noting that the necessity of negative edges has been highlighted in~\citet{fagcn} and~\citet{gprgnn} using the graph filtering concept. \citet{fagcn} use attention mechanism to learn negative edges. On the other hand, \citet{gprgnn} uses a polynomial function with negative weights to obtain negative edges. 

\subsubsection{\label{sec:EigenEigen}\textbf{Hard Low Pass (HLP) Aggregation Model}}
It is often useful to reduce dimension of raw features using principal component analysis. Let  ${\mathbf Q} \adjS_x {\mathbf Q}^T$ be the TSVD of $\feat\feat^T$.  
upon substituting node embedding with TSVD of raw features (\ref{eqn:truncatedEigen}), we get: 
\begin{equation}
    {\mathbf S} = \eigenU (\adjS_{k_{1}}^{A}) \eigenU_{\tilde x} (\adjS_{k_{2}}^{X})
\label{eqn:eigeignn}
\end{equation}
where $\eigenU_{\tilde x} = \eigenU^T {\mathbf Q}$. We refer (\ref{eqn:eigeignn}) as HLP Aggregation model as it involves truncated singular vectors of both $\adj$ and $\feat$. We treat $k_{1}$ and $k_{2}$ as hyper-parameters.  

\subsubsection{\textbf{Hard Low Pass (HLP) Concat Model}} 
We suggest a simple alternative modeling approach that works quite well for heterophilic graphs. With neighborhood aggregation, difficulties arise when ${\adj}$ and ${\mathbf X}$ are \textit{incompatible} in the sense that  it degrades the performance due to a violation of assumptions made. Though graph adaptation methods try to mitigate the effect of any violation, they still operate within the field of improving neighborhood aggregation. Therefore, it may be difficult to improve beyond some limits with the neighborhood aggregation restriction. Also, it may only add more computational burden. In this context, we explored the approach of concatenating truncated node features (${\mathbf X})$ and truncated eigenvectors of $\adj$, and learning a classifier model. Since the features are decoupled now, this model is less affected by the incompatibility between $\adj$ and $\feat$. Additionally, there is a significant reduction in computational cost. Therefore, the HLP Concat model is faster to train. We found this simple approach to outperform state-of-the-art methods on several heterophilic benchmark datasets. 

In the following section, we discuss our experimental section and results.


%% file: sections/experiments.tex
\section{Experiments}
\label{sec:experiments}
We validate our proposed models by comparing against several baselines and state-of-the-art heterophily graph networks on node classification task.

\subsection{Datasets}
We evaluate on seven heterophilic datasets to show the effectiveness of our models. Detailed statistics of the datasets used are provided in Table~\ref{tab:stats_table}. We borrowed \textbf{Texas}, \textbf{Cornell}, \textbf{Wisconsin} from WebKB\footnote{http://www.cs.cmu.edu/afs/cs.cmu.edu/project/theo-11/www/wwkb}, where nodes represent web pages and edges denote hyperlinks between them. \textbf{Actor} is a co-occurence network borrowed from~\cite{actordataset}, where nodes correspond to an actor, and and edge represents the co-occurrence on the same Wikipedia page. \textbf{Chameleon}, \textbf{Squirrel}, and \textbf{Crocodile} are borrowed from~\cite{chameleondataset}. Nodes correspond to web pages and edges capture mutual links between pages. For all benchmark datasets, we use feature vectors, class labels from~\cite{supergat}. For datasets in (Texas, Wisconsin, Cornell, Chameleon, Squirrel, Actor), we use 10 random splits (48\%/32\%/20\% of nodes for train/validation/test set) from~\cite{geomgcn}. For Crocodile, we create 10 random splits following~\cite{supergat}.

\subsection{Methods of Comparison}
We provide the methods in comparison along with the hyper-parameters ranges for each model. For all the models, we sweep the common hyper-parameters in same ranges. Learning rate is swept over [0.001, 0.003, 0.005, 0.008, 0.01], dropout over [0.2, 0.3, 0.4, 0.5, 0.6, 0.7, 0.8], weight decay over [1e-4, 5e-4, 1e-3, 5e-3, 1e-2, 5e-2, 1e-1], and hidden dimensions over [16, 32, 64]. For model specific hyper-parameters, we tune over author prescribed ranges. We use undirected graphs with symmetric normalization for all graph networks in comparison. For all models, test accuracy is reported for the configuration that achieves the highest validation accuracy. We report standard deviation wherever applicable. 

\textbf{LR and MLP:} We trained Logistic Regression classifier and Multi Layer Perceptron on the given node features. For MLP, we limit the number of hidden layers to one. 

\textbf{\sgcn:} \sgcn~\citep{sgcn} is a spectral method that models a low pass filter and uses a linear classifier. The number of layers in \sgcn~is treated as a hyper-parameter and swept over [1, 2].  

\textbf{\supergat:} \supergat~\citep{supergat} is an improved graph attention model designed to also work with noisy graphs. \supergat~ employs a link-prediction based self-supervised task to learn attention on edges. As suggested by the authors, on datasets with homophily levels lower than 0.2 we use \supergat\textsubscript{SD}. For other datasets, we use \supergat\textsubscript{MX}. We rely on authors code\footnote{https://github.com/dongkwan-kim/SuperGAT} for our experiments.

\textbf{\geomgcn:} \geomgcn~\citep{geomgcn} proposes a geometric aggregation scheme that can capture structural information of nodes in neighborhoods and also capture long range dependencies. We quote author reported numbers for Geom-GCN. We could not run Geom-GCN on other benchmark datasets because of the unavailability of a pre-processing function that is not publicly available.

\textbf{\hhgcn:} \hhgcn~\citep{h2gcn} proposes an architecture, specially for heterophilic settings, that incorporates three design choices: i) ego and neighbor-embedding separation, higher-order neighborhoods, and combining intermediate representations.  We quote author reported numbers where available, and sweep over author prescribed hyper-parameters for reporting results on the rest datasets. We rely on author's code\footnote{https://github.com/GemsLab/H2GCN} for our experiments.

\textbf{\fagcn:} \fagcn~\citep{fagcn} adaptively aggregates different low-frequency and high-frequency signals from neighbors belonging to same and different classes to learn better node representations. We rely on author's code\footnote{https://github.com/bdy9527/FAGCN} for our experiments.

\textbf{\appnp:} \appnp~\citep{appnp} is an improved message propagation scheme derived from personalized PageRank. \appnp's addition of probability of teleporting back to root node permits it to use more propagation steps without oversmoothing. We use \gprgnn's~ implementation of \appnp~for our experiments.

\textbf{\gprgnn:} \gprgnn~\citep{gprgnn} adaptively learns weights to jointly optimize node representations and the level of information to be extracted from graph topology. We rely on author's code\footnote{https://github.com/jianhao2016/GPRGNN} for our experiments.

\textbf{HLP Models:} We sweep $k_{1}$ in  [1, min(\#nodes, 2048)] and $k_{2}$ in [1, min(\#features, 2048)]. HLP models use a one layer MLP as the classifier. We restrict to 2048 dimensions to reduce the computation burden. Unlike aggregation models, HLP concat. model need not be restricted to symmetric normalization. Hence, for HLP Concat. model, we sweep the graph type in [directed-graph, undirected-graph], and graph norm type in [no-norm, row-norm, sym-norm].

All models use the Adam optimizer~\cite{adam}. For our proposed models that involve learning, we set early stopping to 30 and maximum number of epochs to 300. We utilize learning rate with decay, with decay factor set to 0.99 and decay frequency set to 50. All our experiments were performed on a machine with Intel Xeon 2.60Ghz processor, 112GB Ram, Nvidia Tesla P-100 GPU with 16GB of memory, python 3.6, and Tensorflow 1.15\citep{tensorflow}. We used Optuna \citep{optuna} to optimize the hyperparameter search. 

\subsection{Experimental Results}

We propose two models HLP Agg. and HLP Concat. Our simple aggregation-based model outperforms popular message-passing approaches like \gcn~ and \sgcn. As opposed to modified aggregation schemes proposed in \citet{supergat, geomgcn}, our simple aggregation based on truncation tends to be more effective for heterophilic settings. 

The disparity between node features and graphs can affect the performance of aggregation schemes. To decouple this effect, we propose concatenation-based methods that effectively leverage signals from the given topology and features. We observe from Table~\ref{tab:main_results} that our proposed concatenation-based method outperforms state-of-the-art approaches on benchmark datasets. We see massive performance gains of up to $\sim$ 30\%.

\fagcn, \gprgnn~have identified that high-frequency components are beneficial for improving task performance on heterophilic graphs. However, our results suggest that carefully selecting low-frequency components can lead to significant performance gains. We believe that the negative edges induced by TSVD are capturing high-frequency components and leading to improvements in performance, similar to how high-frequency coefficients were inducing negative edge-weights in \fagcn. 

\subsection{Ablation Study}
\subsubsection{\textbf{Effect of varying Truncated SVD dimensions}}
We observe in Table~\ref{tab:var_k} that varying dimensions as opposed to fixing dimensions for node features and graphs in general lead to significant improvements in performance. For instance, performance on the Squirrel dataset jumps from 56.59\% to 74.17\%. We believe that, by varying $k_{1}$, the optimal set of negative edges induced can be found that improves performance.

\subsubsection{\textbf{Effect of varying graph normalization}}
In Table~\ref{tab:graph_norm}, we observe for the HLP Agg. model that varying graph normalization over [no-norm., row-norm., symmetric-norm.] set, we see significant gains in performance. We conjecture that the usual symmetric normalization might be limiting aggregation-based methods' performance. However, moving to other normalization schemes will require rethinking the theoretical properties. This is beyond the scope of this work.

\subsubsection{\textbf{TSNE Plots}} 
\begin{figure}%
    \centering
    \subfigure[\gprgnn]{
        \label{fig: GPRGNN_embed}
        \includegraphics[width=0.4\textwidth]{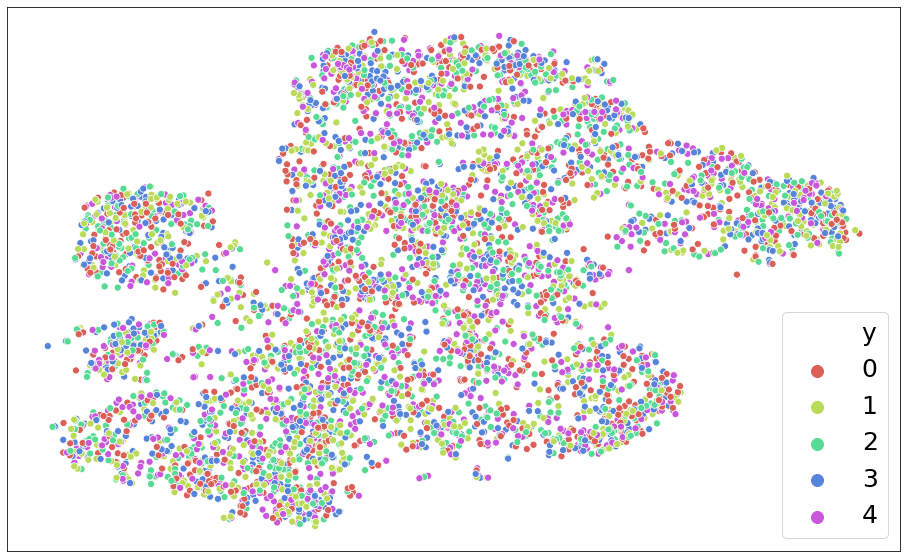}
    }
    \subfigure[HLP Concat]{
        \label{fig:OurModel_embed}
        \includegraphics[width=0.4\textwidth]{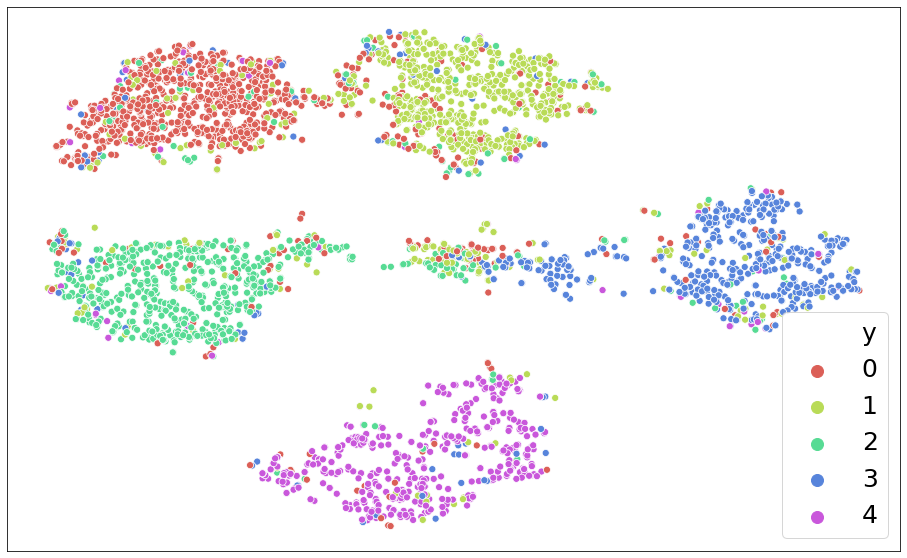}
    }
    \vspace{-5mm}
    \caption{TSNE plots of output layer embedding for Squirrel Dataset}
    \vspace{-5mm}
    \label{fig:embedPlots}
\end{figure}

Figure~\ref{fig:embedPlots} shows the TSNE plots of learned embeddings for \gprgnn~ and HLP Concat model on the Squirrel dataset. We can observe discernible clusters in our learned embedding plot. The plot qualitatively depicts the superiority of our proposed approach. 

\begin{table*}[]
\resizebox{\textwidth}{!}{%
\begin{tabular}{@{}cccccccc@{}}
\toprule
\multicolumn{1}{c|}{} &
  \textbf{Texas} &
  \textbf{Wisconsin} &
  \textbf{Actor} &
  \textbf{Squirrel} &
  \textbf{Chameleon} &
  \textbf{Crocodile} &
  \textbf{Cornell} \\ \midrule
\multicolumn{1}{c|}{\textbf{LR}} &
  81.35 (6.33) &
  84.12 (4.25) &
  34.70 (0.89) &
  34.73 (1.39) &
  45.68 (2.52) &
  53.01 (1.77) &
  83.24 (5.64) \\
\multicolumn{1}{c|}{\textbf{MLP}} &
  81.24 (6.35) &
  84.43 (5.36) &
  \textbf{36.06 (1.11)} &
  35.38 (1.38) &
  51.64 (1.89) &
  54.47 (1.99) &
  83.78 (5.80) \\
\multicolumn{1}{c|}{\textbf{SGCN}} &
  62.43 (4.43) &
  55.69 (3.53) &
  30.44 (0.91) &
  45.72 (1.55) &
  60.77 (2.11) &
  51.54 (1.47) &
  62.43 (4.90) \\
\multicolumn{1}{c|}{\textbf{GCN}} &
  61.62 (6.14) &
  53.53 (4.73) &
  30.32 (1.05) &
  46.04 (1.61) &
  61.43 (2.70) &
  52.34 (2.61) &
  62.97 (5.41) \\
\multicolumn{1}{c|}{\textbf{SuperGAT}} &
  61.08 (4.97) &
  56.47 (3.90) &
  29.32 (1.00) &
  31.84 (1.26) &
  43.22 (1.71) &
  52.41 (1.92) &
  57.30 (8.53) \\
\multicolumn{1}{c|}{\textbf{Geom-GCN}} &
  67.57* &
  64.12* &
  31.63* &
  38.14* &
  60.90* &
  NA &
  60.81* \\
\multicolumn{1}{c|}{\textbf{H2GCN}} &
  84.86 (6.77)* &
  \textbf{86.67 (4.69)*} &
  35.86 (1.03)* &
  37.90 (2.02)* &
  58.40 (2.77) &
  53.17 (1.21) &
  82.16   (4.80)* \\
\multicolumn{1}{c|}{\textbf{FAGCN}} &
  82.43 (6.89) &
  82.94   (7.95) &
  34.87   (1.25) &
  42.59   (0.79) &
  55.22 (3.19) &
  54.35 (1.05) &
  79.19   (9.79) \\
\multicolumn{1}{c|}{\textbf{APPNP}} &
  81.89 (5.85) &
  85.49 (4.45) &
  35.93 (1.04) &
  39.15 (1.88) &
  47.79 (2.35) &
  53.13 (1.93) &
  81.89 (6.25) \\
\multicolumn{1}{c|}{\textbf{GPR-GNN}} &
  81.35 (5.32) &
  82.55 (6.23) &
  35.16 (0.90) &
  46.31 (2.46) &
  62.59 (2.04) &
  52.71 (1.84) &
  78.11 (6.55) \\ \midrule
\multicolumn{1}{c|}{\textbf{HLP Aggregation}} &
  67.57   (4.68) &
  65.49   (3.94) &
  27.78   (0.98) &
  56.59   (1.36) &
  66.36   (2.07) &
  54.58   (1.88) &
  66.76   (6.84) \\ 
\multicolumn{1}{c|}{\textbf{HLP Concat}} &
  \textbf{87.57 (5.44)} &
  \textbf{86.67 (4.22)} &
  34.59 (1.32) &
  \textbf{74.17 (1.83)} &
  \textbf{77.48 (0.80)} &
  \textbf{55.87 (1.25)} &
  \textbf{84.05 (4.67)} \\ \bottomrule
\end{tabular}%
}
\caption{Comparison With Baselines. The results marked with "*" are obtained from the corresponding paper.}
\vspace{-5mm}
\label{tab:main_results}
\end{table*}

\begin{table*}[]
\resizebox{\textwidth}{!}{%
\begin{tabular}{@{}c|ccccccc@{}}
\toprule
\textbf{HLP Concat} &
  \textbf{Texas} &
  \textbf{Wisconsin} &
  \textbf{Actor} &
  \textbf{Squirrel} &
  \textbf{Chameleon} &
  \textbf{Crocodile} &
  \textbf{Cornell} \\ \midrule
\textbf{Fixed \# of Dims.} &
  82.43 (6.97) &
  80.59 (4.42) &
  32.84 (1.34) &
  62.29 (1.50) &
  69.52 (0.87) &
  54.24 (2.67) &
  78.65 (6.67) \\
\textbf{Variable \# of Dims.} &
  \textbf{87.57 (5.44)} &
  \textbf{86.67 (4.22)} &
  \textbf{34.59 (1.32)} &
  \textbf{74.17 (1.83)} &
  \textbf{77.48 (0.80)} &
  \textbf{55.87 (1.25)} &
  \textbf{84.05 (4.67)} \\ \bottomrule
\end{tabular}%
}
\caption{Effect of Varying dimensions for Truncated SVD based features}
\vspace{-5mm}
\label{tab:var_k}
\end{table*}

\begin{table*}[]
\resizebox{\textwidth}{!}{%
\begin{tabular}{@{}c|ccccccc@{}}
\toprule
 \textbf{HLP Aggregation} &
  \textbf{Texas} &
  \textbf{Wisconsin} &
  \textbf{Actor} &
  \textbf{Squirrel} &
  \textbf{Chameleon} &
  \textbf{Crocodile} &
  \textbf{Cornell} \\ \midrule
\textbf{Symmetric Norm} &
  67.57   (4.68) &
  65.49   (3.94) &
  \textbf{27.78   (0.98)} &
  56.59   (1.36) &
  66.36   (2.07) &
  \textbf{54.58   (1.88)} &
  66.76   (6.84) \\
\textbf{Treated as hyperparameter} &
  \textbf{68.65   (7.94)} &
  \textbf{65.69   (3.07)} &
  27.51   (0.83) &
  \textbf{73.62   (1.94)} &
  \textbf{76.51   (1.81)} &
  53.48   (2.30) &
  \textbf{69.46   (4.02)} \\ \bottomrule
\end{tabular}%
}
\caption{Effect of Graph Normalization. When treated as a hyperparameter, we vary the normalization in [symmetric-normalization, row-normalization, no-normalization], tried both the original directed graphs as well as the undirected variant. We pick the best model based on Validation Accuracy.}
\vspace{-5mm}
\label{tab:graph_norm}
\end{table*}

%% file: sections/conclusion.tex
\section{Conclusion}
\label{sec:conclusion}
In this paper, we presented a TSVD based approach inspired by the \gprgnn~\cite{gprgnn} model, which we show can be interpreted as selecting/weighing the singular vectors by scaling the corresponding singular values. We propose a TSVD based regularization model, that enables our model to avoid overfitting and acts as a hard low-pass filter. We show that our models outperform baselines across all heterophilic datasets. This model is simple and computationally cheaper. It begs the question of whether there are alternative ways to model Graph Neural Networks that work across varying homophily scores. We leave it as future work.